\documentclass[letterpaper]{article}
\usepackage{aaai25}

\title{SDOF: Taming the Alignment Tax in Multi-Agent Orchestration\\with State-Constrained Dispatch}

\author{
  Zhantao Wang\\
  Digital China\\
  \texttt{wangztq@digitalchina.com}
}

\date{}

\begin{document}
\maketitle

\begin{abstract}
Multi-agent orchestration frameworks such as LangChain, LangGraph, and CrewAI route tasks through graph-based pipelines but do not enforce the stage constraints that govern real business processes. We present \textbf{SDOF}, a framework that treats multi-agent execution as a constrained state machine. SDOF operates through two primary defensive layers, implemented by three components: (1) an \textbf{Online-RLHF Specialized Intent Router} trained via Generative Reward Modeling (GRPO) and (2) a \textbf{StateAwareDispatcher} with GoalStage finite-automaton checks and precondition/postcondition SkillRegistry validation for auditable execution control. On a recruitment system backed by the Beisen iTalent platform (6,000+ enterprises), 185 expert-curated scenarios trigger \textbf{1,671 live API calls}. Our GSPO-aligned 7B Intent Router achieves higher joint accuracy than zero-shot GPT-4o on this FSM-constrained adversarial routing benchmark (80.9\% vs 48.9\%). In end-to-end execution, SDOF reaches \textbf{86.5\% task completion} (95\% CI 80.8--90.7) and blocks all 22 operations in the injected-illegal HR subset. Under a broader message-level blocking audit, SDOF attains precision 100\% and recall 88\% (expert agreement $\kappa{=}0.94$). A separate evaluation on \textbf{960 SGD-derived dialogues} spanning 8 service domains surfaces 201 stage-order conflicts under our FSM mapping, 41 of which arise in the normal split. This arXiv version reports the current validated scope; extended multi-seed training comparisons and deeper-workflow evaluations will be released in a subsequent update.
\end{abstract}

\section{Introduction}

When LLM-based agents automate enterprise workflows, they must respect the sequential stage constraints that govern each process. In recruitment, for instance, a candidate cannot be evaluated before resume screening, and no offer may be issued before the interview loop concludes. Violations cause compliance failures, data corruption, and legal risk. These constraints differ from generic task dependencies: they are \textit{domain-specific}, \textit{stage-ordered}, and must be enforced at the orchestration layer rather than inside individual agents~\cite{vanderaalst2012process}.

Current orchestration stacks---LangChain~\cite{langchain2023}, LangGraph~\cite{langgraph2024}, CrewAI~\cite{crewai2024}, AutoGen~\cite{wu2024autogen}---excel at routing messages between agents and tools, yet none of them natively checks whether the \textit{current workflow stage} permits a requested action. An agent in the SOURCING phase can therefore call the interview-scheduling API if a graph edge exists, even though the business process forbids it. In regulated industries this gap is unacceptable.

We propose \textbf{SDOF (State-Driven Orchestration Framework)}, which models multi-agent task execution as a constrained state machine with two defensive layers implemented through three main structural additions:

\begin{enumerate}[nosep]
    \item \textbf{Online-RLHF Specialized Intent Router}: A 7B model specialized with online programmatic rewards in veRL (GRPO), achieving higher joint accuracy than GPT-4o zero-shot on our FSM-constrained benchmark.
    \item \textbf{GoalStage FSM \& SkillRegistry}: Intent-stage constraints ($\Lambda$) and precondition validation ($\Pi_{pre}$) for out-of-order risk reduction.
    \item \textbf{StateAwareDispatcher}: Orchestrates execution with stage-filtered skill selection (Algorithm~\ref{alg:dispatch}), enforcing constraints \textit{before} skill binding, and generating replayable audit trails.
\end{enumerate}

We evaluate SDOF on a production-grade intelligent recruitment system integrated with Beisen iTalent (serving 6,000+ enterprises, 48 real job positions), using 185 expert-curated scenarios with 882 messages and 1,671 real API calls. We validate cross-domain generalization on 960 SGD-derived dialogues (800 normal-split + 160 adversarial) across 8 domains~\cite{rastogi2020sgd}.

Our main contributions are:
\begin{itemize}[nosep]
    \item A design contribution: an intent-stage binding formulation ($\Lambda$) that adds an orthogonal constraint layer on top of transition validation.
    \item The SDOF framework itself, packaging GoalStage FSM, SkillRegistry, and StateAwareDispatcher into a reusable orchestration layer.
    \item An evaluation spanning two data sources---185 HR scenarios with live Beisen API calls and 960 SGD-derived dialogues across 8 service domains---plus expert validation ($\kappa{=}0.94$).
\end{itemize}

\section{Related Work}

\textbf{LLM Agent Orchestration.}
LangChain~\cite{langchain2023} popularized chaining LLM calls with tool invocations. LangGraph~\cite{langgraph2024} layers a directed graph over this, allowing cyclic agent interactions and persistent state. CrewAI~\cite{crewai2024} organizes agents into role-based teams that delegate hierarchically. AutoGen~\cite{wu2024autogen} takes a different angle, letting multiple agents converse in group-chat topologies. MetaGPT~\cite{hong2024metagpt} is closest in spirit to our work: it structures agent collaboration around Standard Operating Procedures (SOPs), but its SOPs are rigid sequences rather than constraint-based state machines with precondition checks. AgentScope~\cite{gao2024agentscope} targets distributed deployment without stage-level enforcement; AgentVerse~\cite{chen2023agentverse} focuses on emergent behavior rather than formal workflow guarantees. For broader context, Wang et al.~\cite{wang2023survey} and Xi et al.~\cite{xi2023rise} survey the space.

These frameworks expose different orchestration primitives: LangGraph centers on transition graphs, AutoGen on group-chat / selector / swarm teams, and MetaGPT on SOP-driven role teams. However, in their native forms they do not expose \textit{business-stage legality} as an explicit runtime contract of the kind evaluated here. Table~\ref{tab:framework_caps} summarizes the practical distinction.

Recent harness-style runtimes further expand the design space in a different direction: they package long-running execution features such as middleware-managed delegation, sandboxing, memory injection, summarization, scheduling, and operator intervention into general-purpose agent runtimes. These systems are useful evidence that agent quality increasingly depends on runtime organization outside the base model. Their design emphasis, however, is usually task continuity and operational breadth rather than \textit{workflow legality under an explicit business FSM}. SDOF therefore occupies a narrower but practically distinct niche.

\begin{table}[t]
\centering
\caption{Capability-level comparison of representative orchestration frameworks.}
\label{tab:framework_caps}
\scriptsize
\setlength{\tabcolsep}{3pt}
\resizebox{\columnwidth}{!}{%
\begin{tabular}{@{}lp{0.21\columnwidth}p{0.19\columnwidth}p{0.19\columnwidth}p{0.14\columnwidth}@{}}
\toprule
\textbf{Framework} & \textbf{Native Primitive} & \textbf{Stage Legality} & \textbf{Pre-exec Guard} & \textbf{Auditable State} \\
\midrule
LangGraph & State graph & Transition graph only & External/manual & External \\
AutoGen & Group chat / GraphFlow & Not native & Custom wrapper & External \\
MetaGPT & SOP / role team & SOP discipline & Prompt/action checks & External \\
SDOF & FSM + dispatcher & Yes ($\Lambda$) & Yes ($\Pi_{pre}$) & Audited \\
\bottomrule
\end{tabular}
}
\end{table}

\textbf{Tool Use and API Integration.}
A separate line of work concentrates on \textit{how} LLMs invoke external tools. Toolformer~\cite{schick2023toolformer} lets models learn tool calls from self-supervision; ReAct~\cite{yao2022react} interleaves chain-of-thought reasoning with action execution; Reflexion~\cite{shinn2023reflexion} introduces verbal self-critique after failures. At the API level, ToolLLM~\cite{qin2023toolllm} benchmarks 16,000+ real endpoints, RestGPT~\cite{song2023restgpt} targets RESTful services, and Gorilla~\cite{patil2023gorilla} improves API-call accuracy via retrieval. All of these address the \textit{capability} to call tools---none address \textit{when} a call is legally permitted given a workflow's current stage.

\textbf{State Machines for LLM Agents.}
StateFlow~\cite{wu2024stateflow} maps LLM task-solving onto finite state machines to structure intermediate steps. TaskWeaver~\cite{during2024taskweaver} adopts a code-first planning style; DSPy~\cite{khattab2023dspy} compiles declarative LLM programs into optimized pipelines. These systems impose structure on \textit{computation} but not on \textit{business-level stage legality}.

\textbf{Planning as an Externalized Systems Capability.}
An emerging line of agent engineering externalizes planning from latent chain-of-thought into explicit system structures: plan artifacts, runtime todo state, delegated planner/executor roles, middleware-enforced checkpoints, and evaluation harnesses. This perspective is useful for positioning SDOF. We do not claim a universal planner; rather, SDOF externalizes one enterprise-relevant slice of planning---\textit{whether an action is legally executable at the current workflow stage}---into an auditable orchestration contract.

\textbf{Safe and Constrained LLM Systems.}
Guardrail methods filter model \textit{outputs}---harmful tokens, hallucinated facts, policy violations~\cite{dafoe2021cooperative}. SDOF operates one layer earlier: it prevents \textit{actions} that violate the process model before any agent executes them, analogous to compile-time type checking versus runtime assertions. Recent safety benchmarks ASSEBench~\cite{assebench2025} and AMA-Bench~\cite{amabench2026} report severe failures on context-dependent privilege escalation---a failure mode SDOF's two-layer FSM+precondition architecture is designed to mitigate. The process-mining literature~\cite{vanderaalst2012process} offers techniques for discovering FSM-like workflow models from logs, a direction that could automate SDOF's currently manual stage definitions.

\textbf{Agent Memory Mechanisms.}
Recent benchmarks expose a growing gap between agent memory capability and real-world demands. LoCoMo~\cite{locomo2024} reveals that LLM agents fail on long-horizon temporal causal reasoning across sessions---the same class of failures SDOF's GoalStage FSM addresses by making session state explicit and persistent. MemoryArena~\cite{memoryarena2026} demonstrates that individual agent memory fails in interdependent multi-agent tasks where shared state consistency is critical. Our GoalManager (backed by PostgreSQL) represents a concrete instantiation of the \textit{shared procedural memory} primitive these benchmarks call for.

More broadly, most prior memory work treats memory as a retrieval or continuity mechanism. SDOF instead uses memory as an \textit{active governance substrate}: workflow state is scoped by \texttt{goal\_id}, mutated only through stage-legal transitions, and mirrored by replayable \texttt{ProcessEvent} traces. In enterprise settings, remembering facts is insufficient; the system must also remember which workflow owns the state, who is allowed to advance it, and under which preconditions the change is auditable.

\textbf{Key gap.} No prior framework simultaneously provides (1)~intent-stage binding that is orthogonal to transition graphs, (2)~precondition validation at the skill level, (3)~evaluation against live production APIs, and (4)~a persistent shared memory substrate for multi-agent coordination. MetaGPT~\cite{hong2024metagpt} comes closest but omits the two-layer check (stage $+$ precondition) and has no real-API validation. Our focus is orthogonal to communication-topology optimization: regardless of whether agents are connected through a fixed graph, a supervisor, or a learned routing policy, SDOF constrains which actions are legal at the orchestration layer.

\section{System Architecture}

\subsection{Overview: A Harness Control Architecture}

\begin{figure*}[t]
\centering
\includegraphics[width=0.95\textwidth]{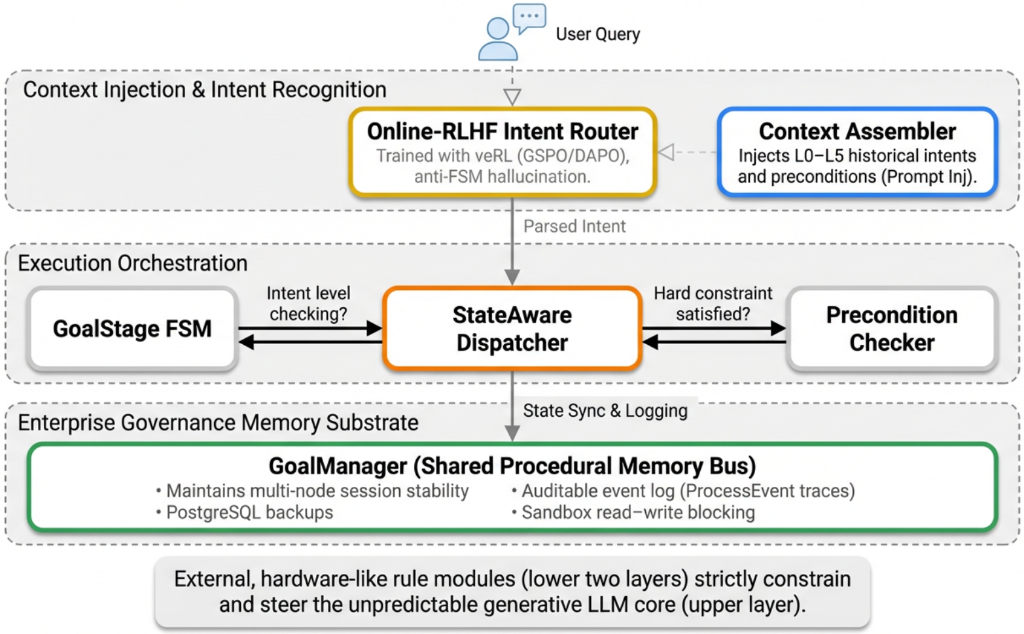}
\caption{SDOF as an enterprise harness architecture. The generative LLM core (top layer) is constrained by deterministic orchestration and memory modules (bottom layers) to reduce unsupported stage transitions and uncontrolled workflow drift.}
\label{fig:arch}
\end{figure*}

Figure~\ref{fig:arch} illustrates the SDOF architecture from a control perspective. Instead of treating the agent as an unbound generative model, SDOF wraps the LLM core within a harness-style architecture. User messages flow through the IntentRouterAgent for intent recognition. This upper layer is then constrained by two external rule-governed modules: the \textit{Execution Orchestration} layer (which checks stage and precondition constraints via the StateAwareDispatcher) and the \textit{Enterprise Governance Memory Substrate}.

Unlike general harness runtimes that optimize for broad task continuity across many open-ended activities, SDOF specializes the runtime around enterprise process legality: state ownership, stage-legal dispatch, precondition-bounded execution, and replayable audit trails. The design goal is not to maximize generic autonomy, but to ensure that each step remains lawful with respect to the business process being automated.

A practical design choice is that GoalManager is not merely a persistence cache for the current stage. It acts as a goal-scoped \textit{governance memory layer} spanning four record classes in the implementation---goal, position, candidate, and process event---so that each dispatch step is bound to a workflow owner (\texttt{goal\_id}), current stage, mutable business state, and replayable audit history. This converts memory from passive conversation storage into an active control surface consulted during dispatch.

\subsection{GoalStage Finite Automaton}

We define the workflow automaton as a tuple $\mathcal{G} = (S, s_0, T, \delta, I, \Lambda)$ where:
\begin{itemize}[nosep]
    \item $S = \{\text{init, src, int, off, onb, close}\}$: workflow stages
    \item $s_0 = \text{init}$: initial stage
    \item $T \subseteq S \times S$: legal transitions
    \item $I$: intent set (create\_demand, screen\_resume, etc.)
    \item $\Lambda: I \to 2^S$: \textbf{intent-stage binding}
\end{itemize}
\textit{Stage semantics.} In this paper, \texttt{init}=initialization, \texttt{src}=sourcing, \texttt{int}=interview, \texttt{off}=offer, \texttt{onb}=onboarding, and \texttt{close}=workflow closure.

\textbf{Definition 1} (Intent-Stage Binding). For each intent $i \in I$, $\Lambda(i) \subseteq S$ defines the stages where $i$ is \textit{legally executable}. An intent $i$ at stage $s$ is \textit{stage-legal} iff $s \in \Lambda(i)$.

\textbf{Practical distinction.} LangGraph defines $T$ (valid transitions) but not $\Lambda$ (intent-stage binding). An agent in SOURCING state can execute \texttt{evaluate\_candidate} if a graph edge to INTERVIEW exists. SDOF requires $s \in \Lambda(i)$, an \textit{orthogonal} constraint.

\subsection{Formal Problem Definition \& Alignment Tax}

To rigorously define the limitations of pure LLM alignment in enterprise workflows, we formalize the agent execution context as a tuple $\mathcal{E} = \langle \mathcal{M}, \Pi, \Phi \rangle$, where $\mathcal{M}$ is the GoalStage automaton, $\Pi$ is the LLM policy, and $\Phi$ represents the structural syntactic preconditions (e.g., JSON schema adherence and parameter constraints).

\textbf{Definition 2} (The Alignment Tax for Structured Tasks). Let $P(\Phi \mid x, \Pi)$ be the probability that policy $\Pi$ generates a structurally valid output given input $x$. A strong reasoning model $\Pi_{think}$ introduces latent chain-of-thought tokens $z \sim \Pi_{think}(\cdot \mid x)$ before generating the final action string $y$. The alignment tax is the structural degradation caused by intermediate reasoning:
$$ \Delta_{tax} = P(\Phi \mid x, \Pi_{base}) - P(\Phi \mid x, z, \Pi_{think}) $$
Empirically, as the trajectory $|z|$ grows, the model over-conditions on semantic reasoning at the expense of rigid syntactical boundaries, leading to $\Delta_{tax} \gg 0$. In this paper, we operationalize $\Delta_{tax}$ with an empirical proxy: the drop in structural-validity rate between matched checkpoints under Think and No-Think decoding (Table~\ref{tab:think_ablation}). SDOF mitigates the downstream risk of such degradation via explicit orchestration checks that block execution when legality conditions fail.

\textbf{Execution rule (informal).} In SDOF, an intent is executed only when $s \in \Lambda(i)$ and $precond(i)=\text{True}$. This rule defines the runtime legality boundary used by the dispatcher.

\begin{figure}[t]
\centering
\includegraphics[width=\columnwidth]{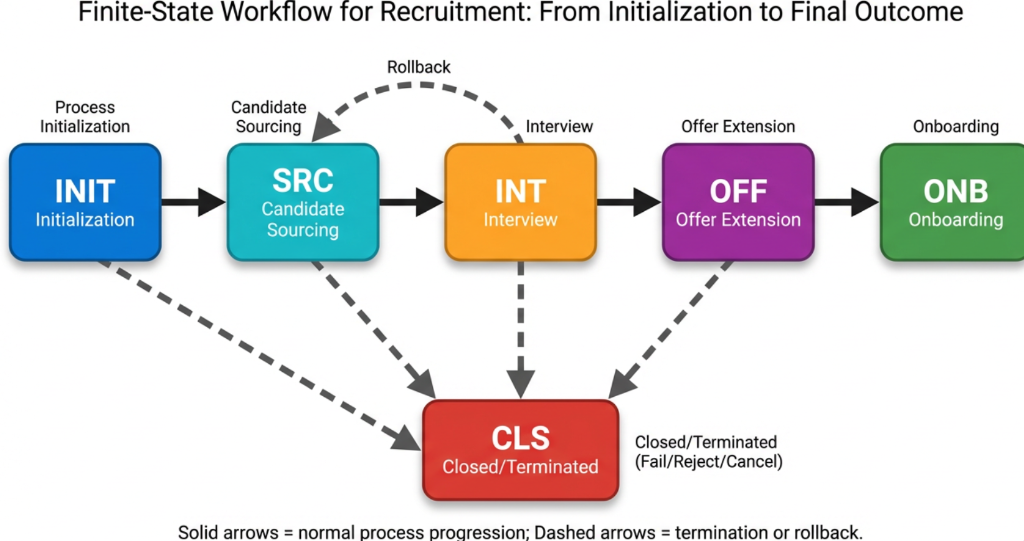}
\caption{Finite-State Workflow (FSM) for Recruitment. Solid arrows indicate normal state progression, while dashed arrows indicate rollback to a previous state or early termination.}
\label{fig:fsm}
\end{figure}

\subsection{SkillRegistry with Formal Preconditions}

\textbf{Definition 3} (Skill Specification). Each skill $sk \in \mathcal{R}$ is a tuple:
$sk = (\text{id}, \ell, \Sigma_{sk}, \Pi_{pre}, \Pi_{post}, \rho)$
where $\ell \in \{L0, L1, L2\}$ is the risk level, $\Sigma_{sk} \subseteq S$ is the set of applicable stages, $\Pi_{pre}$ are preconditions, and $\rho$ is the risk classification.

A skill $sk$ is \textit{precondition-satisfied} in context $\mathcal{C}$ if and only if
\[
\forall \pi \in \Pi_{pre}(sk), \ \pi(\mathcal{C}) = \top.
\]

\begin{table}[t]
\centering
\caption{SkillRegistry three-level classification.}
\label{tab:skills}
\small
\begin{tabular}{@{}cllll@{}}
\toprule
\textbf{Lv} & \textbf{Type} & \textbf{Example} & \textbf{$\Sigma$} & \textbf{$\Pi_{pre}$} \\
\midrule
L0 & Atomic & \texttt{get\_job\_list} & $S$ & $\emptyset$ \\
L1 & Composite & \texttt{pull\_parse} & \{src\} & $\{\pi_1\}$ \\
L1 & Composite & \texttt{screen} & \{src\} & $\{\pi_1, \pi_2\}$ \\
L2 & Policy & \texttt{ask\_missing} & $S$ & $\emptyset$ \\
\bottomrule
\end{tabular}
\end{table}

Beyond stage applicability, each \texttt{SkillSpec} is also governed by disclosure and trust boundaries. In the implementation, low-context L0 manifests are exposed during routing, while richer L1/L2 descriptions are loaded only after a skill is bound; higher-risk or lower-trust skills can therefore be withheld from the candidate set until needed. This progressive disclosure reduces context bloat while aligning capability exposure with enterprise permission boundaries.

\subsection{Safety Properties}

\textbf{Operational safety invariant.} For execution traces $\tau = \langle (m_1, s_1, sk_1), \ldots, (m_n, s_n, sk_n) \rangle$ produced by Algorithm~\ref{alg:dispatch}, the dispatcher enforces stage legality before skill execution and validates transitions before committing state updates.

\textbf{Empirical first-line defense.} In our ablation logs, removing stage checking increases precondition failures: $|B_{\neg\text{stage}}| = 175 \gg |B_{\text{full}}| = 22$.

\subsection{Algorithm: StateAwareDispatch}

\begin{algorithm}[t]
\caption{StateAwareDispatch}
\label{alg:dispatch}
\begin{algorithmic}[1]
\REQUIRE message $m$, context $ctx$, GoalManager $G$, SkillRegistry $\mathcal{R}$
\ENSURE DispatchResult
\STATE $\text{intent} \leftarrow \text{IntentRouter.identify}(m)$
\STATE $sk \leftarrow \mathcal{R}.\text{select\_skill}(\text{intent}, s)$ \COMMENT{stage-filtered}
\IF{$sk = \text{NULL}$}
    \RETURN \textsc{SKILL\_NOT\_FOUND}
\ENDIF
\IF{$\neg \forall p \in sk.\text{pre}: ctx.\text{check}(p)$}
    \STATE $G.\text{log}(\text{PRECONDITION\_FAIL})$
    \RETURN \textsc{PRECONDITION\_FAIL}
\ENDIF
\STATE $\text{result} \leftarrow \text{executor}(sk, ctx)$
\STATE apply postconditions($sk, ctx, \text{result}$)
\STATE $\text{target} \leftarrow \text{StageMap}(\text{intent})$
\IF{target $\neq s$ \AND $s.\text{can\_transition}(\text{target})$}
    \STATE $G.\text{advance\_stage}(s \to \text{target})$
\ELSIF{target $\neq s$}
    \RETURN \textsc{ILLEGAL\_TRANSITION}
\ENDIF
\STATE $G.\text{log}(\text{SUCCESS}, sk, \text{result})$
\RETURN \textsc{SUCCESS}
\end{algorithmic}
\end{algorithm}

Taken together, GoalManager and SkillRegistry instantiate four complementary memory roles: \textit{working memory} (current stage and session variables), \textit{procedural memory} (goal-scoped workflow state and transition history), \textit{reference memory} (skill documentation loaded on demand), and \textit{audit memory} (replayable \texttt{ProcessEvent} traces). The dispatcher queries and updates these memories on every step rather than treating memory as a passive retrieval backend.
In this sense, memory functions as an \textit{active control interface} rather than a passive recall layer: the dispatcher consults memory to decide whether a transition is legal, which preconditions remain unsatisfied, and how the workflow should be explained to a human operator.

\section{Implementation}

The system is deployed as an intelligent recruitment assistant integrated with the Beisen iTalent platform, which serves over 6,000 enterprises across China.

\subsection{Agent Architecture}

SDOF orchestrates \textbf{7 specialized agents}, each responsible for a specific recruitment function:
\begin{itemize}[nosep]
    \item \textbf{Job Requirement Agent}: Creates and manages job postings via Beisen API
    \item \textbf{Resume Screening Agent}: Pulls candidates from talent pool and applies screening criteria
    \item \textbf{Candidate Invitation Agent}: Manages interview scheduling and notifications
    \item \textbf{Interview Questions Agent}: Generates role-specific interview questions
    \item \textbf{Interview Rounds Agent}: Configures multi-round interview structures
    \item \textbf{Interview Evaluation Agent}: Collects and aggregates interviewer feedback
    \item \textbf{Interview Summary Agent}: Produces comprehensive candidate reports
\end{itemize}

\subsection{API Integration}

The system connects to Beisen iTalent via OAuth2-authenticated REST APIs (tenant ID: 430008). The production environment contains 48 real job positions. During evaluation, the system invokes \texttt{GetJobList}, \texttt{GetApplicantList}, and \texttt{DispatchAction} endpoints totaling 1,671 real API calls.

\subsection{Skill Configuration}

The SkillRegistry contains \textbf{10 registered skills}: 4 L0 (atomic queries, universally available), 4 L1 (composite operations with stage and precondition constraints), and 2 L2 (policy-level fallback handlers). Intent recognition uses a \textbf{dual-mode} approach: string-match (0.12ms, 97.5\% STA, where STA denotes State Transition Accuracy) for deterministic intents, with LLM fallback for ambiguous cases.

\section{Experiments}

\subsection{Setup}

\textbf{Scenarios.} We construct a domain-specific evaluation benchmark through a structured expert-driven process. First, domain specialists with experience in enterprise recruitment systems define scenario templates covering six categories of workflow interactions, including both valid flows and adversarial constraint-violation attempts. These templates are then systematically instantiated with variable combinations to produce 185 test scenarios comprising 882 messages (Table~\ref{tab:scenarios}). Each scenario is reviewed for realism by a domain expert. \textit{While the dialogue construction follows this structured process, all execution frameworks run against real production environments and invoke live APIs.}

\begin{table}[t]
\centering
\caption{Scenario distribution.}
\label{tab:scenarios}
\small
\begin{tabular}{@{}llp{3.5cm}@{}}
\toprule
\textbf{Category} & \textbf{n} & \textbf{Description} \\
\midrule
Normal & 50 & Complete recruitment flows \\
Illegal & 25 & Stage-skipping violations \\
Rollback & 25 & Backward transitions \\
Multi & 25 & Multi-candidate operations \\
Abort & 30 & Early termination \\
Concurrent & 30 & Parallel multi-candidate operations and shared-resource scheduling conflicts \\
\bottomrule
\end{tabular}
\end{table}

\textbf{Baselines.} (1)~Vanilla: no constraints. (2)~\textbf{LangGraph (v1.0.9)}: real \texttt{StateGraph}. (3)~\textbf{LangGraph+Pre}: LangGraph with precondition checking. We select LangGraph as the primary executable baseline because it is the only widely used comparator in our set that natively exposes a transition-graph API directly mappable to the 185-scenario suite. AutoGen provides group-chat / swarm / GraphFlow teams, and MetaGPT provides SOP-driven role teams; both are useful architectural comparators (Table~\ref{tab:framework_caps}) but are not included as like-for-like baselines in this release because they are not instrumented under the same transition-graph, legality-check, and audit protocol used in our shared-suite comparison. A custom legality wrapper would add substantial non-native logic, making attribution ambiguous in this release. More broadly, recent long-running agent runtimes are informative design context for delegation, memory, and scheduling, but they are not like-for-like baselines for \textit{legality-governed business workflow execution}.

\textbf{Metrics.} TCR (Task Completion Rate), STA (State Transition Accuracy), CVR (Constraint Violation Rate), TRC (Traceability Rate), and LAT (Latency). CVR is computed as the fraction of violating dispatch events over all dispatch events in the evaluated split. TRC measures replayable per-step trace coverage and is only directly comparable for methods that emit such traces. `Raw Blk` below reports the number of blocked operations; blocking correctness is evaluated separately in Table~\ref{tab:blocking}.

\textbf{Noise isolation.} To distinguish genuine capability gaps from infrastructure artifacts, we apply three controls: (1)~API endpoints are pinned to a fixed tenant environment (Beisen tenant 430008) with stable schema versions throughout evaluation; (2)~deterministic string-match intent recognition (97.5\% STA) eliminates stochastic LLM variance for the majority of routing decisions; (3)~latency measurements exclude network round-trip jitter by reporting only the dispatcher-internal overhead (stage check $+$ precondition validation $<$1\,ms). For the RLHF experiments, all algorithm comparisons (GRPO/GSPO/DAPO) share the same training checkpoint step (\texttt{global\_step\_300}) and identical evaluation data split ($n{=}47$), ensuring that observed differences reflect alignment methodology rather than data or compute confounds. We report this routing benchmark separately from the 185-scenario framework-comparison suite to avoid mixing optimization and orchestration evaluations.

\subsection{Framework Comparison}

\begin{table}[t]
\centering
\caption{Framework comparison on the shared HR scenario suite (185 scenarios, 882 messages). SDOF production runs include 1,671 live API calls; graph baselines are evaluated under the same scenario and metric protocol without matching live-API execution cost.}
\label{tab:comparison}
\small
\begin{tabular}{@{}lcccccc@{}}
\toprule
\textbf{Method} & \textbf{TCR} & \textbf{CVR} & \textbf{TRC} & \textbf{Obs. Blk} & \textbf{LAT} \\
\midrule
\textbf{SDOF} & \textbf{86.5\%} & \textbf{2.5\%} & \textbf{100\%} & \textbf{22} & 57.4ms \\
LG+Pre$^\dagger$ & 88.1\% & 2.8\% & --- & 25 & 0.8ms \\
LangGraph$^\ast$ & 84.9\% & --- & --- & 0 & 1.1ms \\
Vanilla$^\ast$ & 74.6\% & --- & --- & 0 & 0.1ms \\
\bottomrule
\end{tabular}
\begin{flushleft}
\scriptsize
$^\ast$ Dashes indicate \emph{not directly comparable} metrics: the released baseline files do not provide replayable legality traces under the same audit protocol as SDOF. Zeros in Observed Blk denote that no explicit block event was emitted in the released run logs for this shared suite; they do not imply zero governance risk or zero attempted illegal actions. \\
$^\dagger$ LG+Pre's 2.8\% CVR is retained as the reported value in the released \texttt{comprehensive\_v2} baseline artifact, but its trace format is still not directly comparable to SDOF's replayable audit chain.
\end{flushleft}
\end{table}

\textbf{Key findings:} (1)~SDOF outperforms the Vanilla unconstrained baseline by \textbf{+11.9\%} TCR. (2)~SDOF blocks all 22 operations in the injected-illegal subset. (3)~LangGraph (v1.0.9), without explicit legality wiring, allows all 22 injected illegal operations in this suite. (4)~LangGraph+Pre reaches a similar reported CVR, but depends on manually configured precondition logic and does not provide the same auditable execution contract. For unconstrained baselines, we omit governance metrics that are not directly comparable without replayable legality traces; their behavior is instead summarized by observed explicit block counts and by the separate blocking-quality analysis in Table~\ref{tab:blocking}. Here, an observed block count of zero should be read narrowly as ``no explicit block event emitted'' rather than as ``no governance failure present.'' This table should therefore be read as a shared-suite governance comparison, not as a fully matched end-to-end systems benchmark.

\subsection{Ablation Study}

\begin{table}[t]
\centering
\caption{Ablation study (185 scenarios).}
\label{tab:ablation}
\small
\begin{tabular}{@{}lccccc@{}}
\toprule
\textbf{Config} & \textbf{TCR} & \textbf{CVR} & \textbf{TRC} & \textbf{Blk} \\
\midrule
\textbf{SDOF (Full)} & 86.5\% & 2.5\% & \textbf{100\%} & 22 \\
w/o StageCheck & 86.5\% & \textbf{19.8\%} & 100\% & \textbf{175} \\
w/o Precondition & 84.9\% & 2.2\% & 100\% & 19 \\
w/o Audit & 86.5\% & 2.5\% & 0.0\% & 22 \\
\bottomrule
\end{tabular}
\end{table}

Removing StageCheck causes CVR to jump from 2.5\% to \textbf{19.8\%} (+696\%), with blocked operations increasing from 22 to \textbf{175}. This indicates that StageCheck is the primary source of CVR reduction in the current stack. By contrast, removing precondition checks lowers TCR and reduces the raw number of blocks (22$\rightarrow$19) while leaving CVR numerically close to the full system; this suggests that precondition validation mainly protects a smaller subset of stage-legal but semantically unsafe actions rather than dominating the headline CVR. This ablation isolates dispatcher-side governance; isolating the standalone contribution of the intent router and its interaction effect with dispatcher checks is left to future controlled experiments.

\subsection{Performance by Scenario Type}

\begin{table}[t]
\centering
\caption{SDOF performance by scenario type.}
\label{tab:bytype}
\small
\begin{tabular}{@{}lcccccc@{}}
\toprule
\textbf{Type} & \textbf{n} & \textbf{TCR} & \textbf{CVR} & \textbf{Blk} & \textbf{Vio} & \textbf{PreF} \\
\midrule
Normal & 50 & 82.0\% & 0.0\% & 0 & 0 & 0 \\
Illegal & 25 & 88.0\% & 40.7\% & 22 & 16 & 6 \\
Rollback & 25 & 80.0\% & 0.0\% & 0 & 0 & 0 \\
Multi & 25 & 76.0\% & 0.0\% & 0 & 0 & 0 \\
Abort & 30 & 100\% & 0.0\% & 0 & 0 & 0 \\
Concurrent & 30 & 93.3\% & 0.0\% & 0 & 0 & 0 \\
\bottomrule
\end{tabular}
\end{table}

\subsection{Dispatch Trace Analysis}

Following the emerging practice of \textit{Trace Grading}---evaluating agent behavior at the step level rather than only at the final outcome---we treat each StateAwareDispatcher record as a graded execution trace. The real dispatcher produced 882 trace steps across 185 scenarios: 860 \textsc{success} (97.5\%), 16 \textsc{illegal\_transition} (1.8\%), 6 \textsc{precondition\_fail} (0.7\%). Each trace step records the triggering intent, current GoalStage, matched skill, precondition evaluation result, and outcome classification, forming a complete per-step audit chain. This step-level grading enables fine-grained \textit{error attribution}: rather than reporting only whether a scenario succeeded or failed, we can pinpoint the exact dispatch step where the pipeline diverged from the correct execution path.

\subsection{Case Study: 22 Blocked Operations}

\begin{table}[t]
\centering
\caption{Representative blocked operations (22 total).}
\label{tab:cases}
\scriptsize
\begin{tabular}{@{}clll@{}}
\toprule
\textbf{\#} & \textbf{User Message} & \textbf{Stage} & \textbf{Block} \\
\midrule
1 & Schedule interview & INIT & illegal\_trans \\
2 & Interview feedback & INIT & illegal\_trans \\
3 & Generate test questions & INIT & illegal\_trans \\
4 & Compare candidates & SRC & precond\_fail \\
5 & Invite to interview & INIT & illegal\_trans \\
6 & Re-screen resumes & OFF & illegal\_trans \\
\bottomrule
\end{tabular}
\end{table}

Two defense layers: (1)~\textbf{Stage constraint} (16 cases): intent requires unreached stage. (2)~\textbf{Precondition check} (6 cases): required data not available.

\subsection{API Latency}

\begin{figure}[t]
\centering
\includegraphics[width=\linewidth]{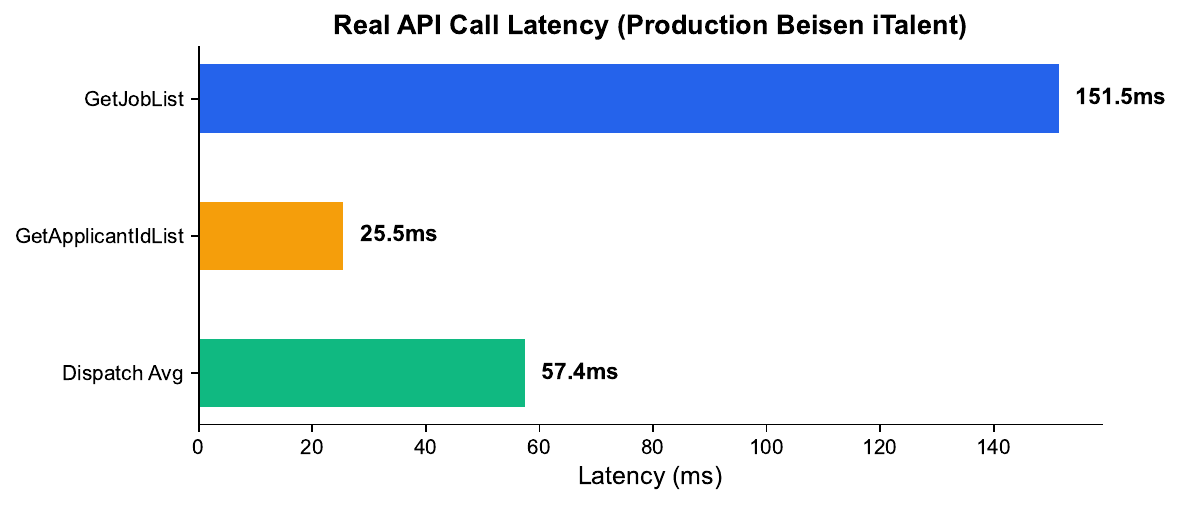}
\caption{Real API call latency measurements.}
\label{fig:latency}
\end{figure}

Average dispatch latency: \textbf{57.4ms} (SDOF production path) vs \textbf{1.1ms} (LangGraph graph-only baseline). The measured gap is dominated by live API latency; LangGraph does not perform matched production API calls, and the legality checks themselves add under 1\,ms.

\subsection{Cross-Domain Generalization (SGD Dataset)}

To validate SDOF's domain-agnosticism, we evaluate on a schema-conformant benchmark derived from the Google SGD dataset~\cite{rastogi2020sgd}. We select 8 domains spanning banking, hospitality, transportation, and entertainment. Each domain maps to a 2--3 stage FSM defined by its service API structure (Table~\ref{tab:sgd_mapping}). We use 100 normal-split dialogues per domain and 20 adversarial illegal variants per domain.

\begin{table}[t]
\centering
\caption{SGD domain-to-FSM mapping. Each domain defines stages based on its service API structure.}
\label{tab:sgd_mapping}
\scriptsize
\begin{tabular}{@{}llc@{}}
\toprule
\textbf{Domain} & \textbf{FSM Stages} & \textbf{$|S|$} \\
\midrule
Banks\_1 & CheckBalance $\to$ TransferMoney & 2 \\
Hotels\_1 & SearchHotel $\to$ ReserveHotel & 2 \\
RentalCars\_1 & GetCarsAvailable $\to$ ReserveCar & 2 \\
Events\_1 & SearchEvent $\to$ BuyTicket & 2 \\
Buses\_1 & SearchBus $\to$ BuyTicket & 2 \\
Homes\_1 & SearchHome $\to$ ReserveHome & 2 \\
Media\_2 & SearchMedia $\to$ PlayMedia & 2 \\
Music\_1 & SearchTrack $\to$ PlayTrack & 2 \\
\bottomrule
\end{tabular}
\end{table}

\begin{table}[t]
\centering
\caption{Cross-domain results on the SGD-derived benchmark (960 dialogues, 1,734 turns).}
\label{tab:crossdomain}
\small
\begin{tabular}{@{}llcccc@{}}
\toprule
\textbf{Domain} & \textbf{Type} & \textbf{n} & \textbf{Turns} & \textbf{Blk} & \textbf{Viol} \\
\midrule
Banks\_1 & normal & 100 & 300 & 0 & 0 \\
Banks\_1 & illegal & 20 & 20 & 20 & 20 \\
Hotels\_1 & normal & 100 & 162 & 38 & 38 \\
Hotels\_1 & illegal & 20 & 20 & 20 & 20 \\
RentalCars\_1 & normal & 100 & 180 & 0 & 0 \\
RentalCars\_1 & illegal & 20 & 20 & 20 & 20 \\
Events\_1 & normal & 100 & 184 & 0 & 0 \\
Events\_1 & illegal & 20 & 20 & 20 & 20 \\
Buses\_1 & normal & 100 & 153 & 0 & 0 \\
Buses\_1 & illegal & 20 & 20 & 20 & 20 \\
Homes\_1 & normal & 100 & 200 & 0 & 0 \\
Homes\_1 & illegal & 20 & 20 & 20 & 20 \\
Media\_2 & normal & 100 & 198 & 0 & 0 \\
Media\_2 & illegal & 20 & 20 & 20 & 20 \\
Music\_1 & normal & 100 & 197 & 3 & 3 \\
Music\_1 & illegal & 20 & 20 & 20 & 20 \\
\midrule
\textbf{Total} & -- & \textbf{960} & \textbf{1,734} & \textbf{201} & 201 \\
\bottomrule
\end{tabular}
\end{table}

\begin{figure}[t]
\centering
\includegraphics[width=\linewidth]{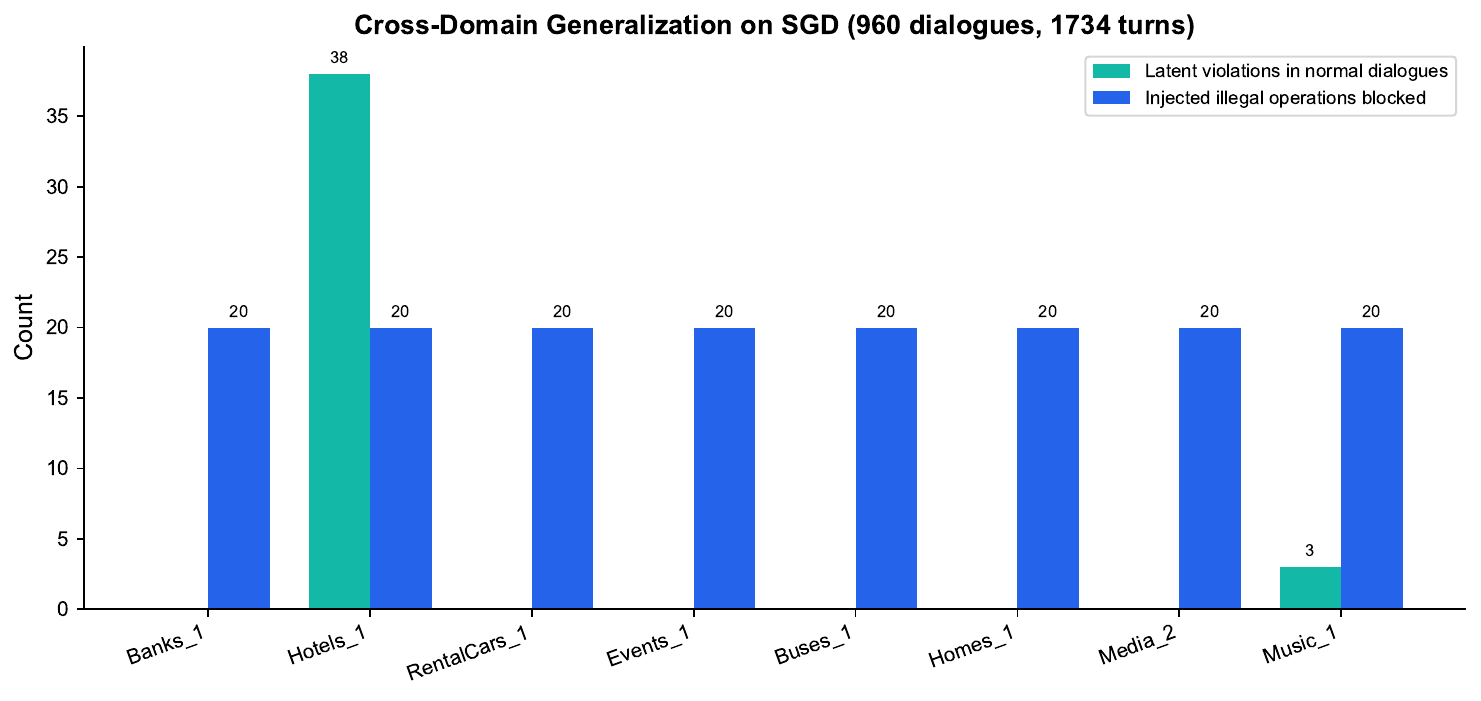}
\caption{Cross-domain generalization on the SGD-derived benchmark (8 domains).}
\label{fig:crossdomain}
\end{figure}

On the 160 injected-illegal messages across 8 domains, SDOF blocks every injected attack. Under the broader message-level blocking evaluation, it attains \textbf{100\% precision (0 false positives)} and 88\% recall. Additionally, 41 latent violations are detected in the normal split. Here, a \textit{latent violation} denotes a request that conflicts with the domain FSM stage order but appears in the non-adversarial portion of the benchmark. In this release, these labels are derived from the same \texttt{expected\_legal} rule-based annotation protocol used in Table~\ref{tab:blocking}; targeted human adjudication of borderline latent cases will be included in a subsequent revision. Hotels\_1 yields the highest latent violation count (38), where users directly request reservations without prior search. Music\_1 shows a similar pattern (3 latent violations), confirming that stage-skipping requests appear even outside the explicitly injected illegal set.

\subsection{Blocking Correctness Evaluation}

Using the released \texttt{expected\_legal} labels over all 882 messages, SDOF achieves \textbf{100\% precision} (0 false positives), 88\% recall, and F1=93.6\%. The 3 false negatives are multi-stage skill availability cases.

\begin{table}[t]
\centering
\caption{Blocking correctness evaluation (882 messages).}
\label{tab:blocking}
\small
\begin{tabular}{@{}lc@{}}
\toprule
\textbf{Metric} & \textbf{Value} \\
\midrule
Accuracy & 99.7\% \\
Precision & \textbf{100.0\%} \\
Recall & 88.0\% \\
F1 Score & 93.6\% \\
True Positive (correctly blocked) & 22 \\
False Positive & 0 \\
False Negative & 3 \\
True Negative & 857 \\
\bottomrule
\end{tabular}
\end{table}

\subsection{Expert Validation of Blocking Decisions}

To validate the correctness of SDOF's blocking decisions beyond algorithmic evaluation, two domain experts independently reviewed all 22 blocked operations and a random sample of 100 permitted operations (122 decisions total). Each annotator labeled whether the blocking decision was \textit{correct}, \textit{incorrect}, or \textit{ambiguous} given the business process constraints.

\begin{table}[t]
\centering
\caption{Expert validation of blocking decisions (122 reviewed).}
\label{tab:humaneval}
\small
\begin{tabular}{@{}lc@{}}
\toprule
\textbf{Metric} & \textbf{Value} \\
\midrule
Annotator agreement (Cohen's $\kappa$) & 0.94 \\
SDOF--Expert agreement & 97.5\% \\
Correctly blocked (of 22) & 22/22 \\
Correctly permitted (of 100) & 97/100 \\
Ambiguous cases & 3 \\
\bottomrule
\end{tabular}
\end{table}

Both experts agreed that all 22 blocked operations were correctly blocked ($\kappa = 0.94$, near-perfect agreement). Three permitted operations were marked ambiguous---cases where multi-stage skills could reasonably be blocked or permitted depending on interpretation. These correspond to the 3 false negatives in Table~\ref{tab:blocking}.

\subsection{Error Analysis}

The 3 false negatives (missed blocks) share a common pattern: multi-stage skills registered as available in multiple stages. For example, \texttt{get\_job\_list} is an L0 skill available in all stages ($\Sigma = S$). When invoked with implicit intent to screen candidates (requiring SRC stage), SDOF permits it because the skill's stage constraint is satisfied. This reveals a limitation in skills with broad stage applicability where the \textit{intent} is contextually stage-specific but the \textit{skill} is not.

\subsection{Intent Router Specialization via Online RLHF}
To ensure the multi-agent orchestration framework is resilient against out-of-order intents (e.g., initiating an interview before a candidate is evaluated), we train the IntentRouterAgent with online reinforcement learning in \textbf{veRL}, comparing \textbf{GRPO} (group-relative policy optimization), \textbf{GSPO}, and \textbf{DAPO} as implemented in that codebase. We model intent parsing as strict constraint satisfaction over the GoalStage FSM with programmatic, zero-tolerance rewards on violations.

\textbf{Implementation Details.} We employ Qwen2.5-7B-Instruct as our policy model. During the GRPO rollout phase, we utilized a group size of $G=2$ responses per prompt, optimized via AdamW with a learning rate of $1\times 10^{-6}$. To heavily penalize stage violations, we computed programmatic zero-tolerance rewards against the FSM rather than relying on a static reward model. The KL penalty coefficient was set to $0.0$ to encourage maximum exploration of the adversarial FSM boundaries. Generation parameters were set to temperature $= 1.0$, top\_p $= 1.0$. The model was trained asynchronously on 8 NVIDIA GPUs.

\begin{table}[htbp]
\centering
\caption{Intent Router Accuracy on Adversarial FSM Sub-split (n=47). Model checkpoints selected via validation metrics.}
\label{tab:intent_router_results}
\resizebox{\columnwidth}{!}{
\begin{tabular}{@{}llccc@{}}
\toprule
\textbf{Model Base} & \textbf{Alignment Method} & \textbf{Intent} & \textbf{Safety} & \textbf{Joint} \\
\midrule
GPT-4o & Zero-shot & 53.2\% & 53.2\% & 48.9\% \\
\midrule
\multicolumn{5}{c}{\textit{Qwen3-8B Base (Deep Reasoning Architecture)}} \\
Qwen3-8B & Zero-shot (think) & 10.6\% & 57.4\% & 6.4\% \\
Qwen3-8B & SFT + GRPO (nothink) & 80.9\% & 57.4\% & 44.7\% \\
Qwen3-8B & SFT + GSPO (nothink) & 78.7\% & 57.4\% & 42.6\% \\
Qwen3-8B & SFT + DAPO (nothink) & 87.2\% & 57.4\% & 48.9\% \\
\midrule
\multicolumn{5}{c}{\textit{Qwen2.5-7B Base (Traditional Foundation)}} \\
Qwen2.5-7B & SFT (Baseline) & 78.7\% & 63.8\% & 51.1\% \\
Qwen2.5-7B & SFT + GRPO & 91.5\% & 78.7\% & 74.5\% \\
Qwen2.5-7B & SFT + DAPO & 91.5\% & 78.7\% & 72.3\% \\
Qwen2.5-7B & \textbf{SFT + GSPO (step=300)} & \textbf{97.9\%} & \textbf{80.9\%} & \textbf{80.9\%} \\
\midrule
\multicolumn{5}{c}{\textit{Scaled Parameter Base (Qwen3-14B, nothink)}} \\
Qwen3-14B & GSPO stageheavy (step=100)$^\ast$ & 85.1\% & 76.6\% & 63.8\% \\
Qwen3-14B & GSPO long-full (step=200)$^\dagger$ & 85.1\% & 76.6\% & 63.8\% \\
\bottomrule
\end{tabular}
}
\vspace{1mm}
\begin{flushleft}
\scriptsize
\textbf{Notes:} Joint Accuracy is computed as the fraction of examples where both intent prediction and safety legality prediction are correct (\texttt{intent\_ok} $\wedge$ \texttt{safety\_ok}). All metrics are on the adversarial test split ($n{=}47$). For fair comparison across algorithms, all RL results (GRPO/GSPO/DAPO) for both Qwen2.5-7B and Qwen3-8B are evaluated at the same training checkpoint (\texttt{global\_step\_300}), as recorded in each \texttt{eval\_*.json} via \texttt{summary.model}. Qwen3 models use \texttt{/nothink} decoding unless stated; see Table~\ref{tab:think_ablation} for think-mode ablation. \\
$^\ast$ For \textit{stageheavy-nothink}, test joint is identical across steps 100--501. We report step=100 for brevity. \\
$^\dagger$ Selected by max validation joint accuracy; \textit{long-full} val\_joint=65.96\% at steps \{200,250,300\} (tie). Test results at step=200.
\end{flushleft}
\end{table}

\textbf{Results and Discussion.} Table~\ref{tab:intent_router_results} presents the evaluation of proprietary baselines against our Online veRL alignments. GPT-4o zero-shot achieves 48.9\% Joint Accuracy, demonstrating vulnerability to implied state transitions caused by conversational \textit{over-helpfulness}. We note that the comparison is deliberately asymmetric: Online RLHF (GRPO/GSPO) requires white-box gradient access to compute programmatic FSM rewards, which is unavailable for proprietary models. GPT-4o therefore serves as a reference point for what a strong closed-source model achieves without domain-specific alignment---not as a like-for-like baseline. We selected the Qwen series (2.5 and 3) as our primary open-weight experimentation chassis to measure alignment behavior across architectures. Our GSPO-aligned Qwen2.5-7B model achieves \textbf{80.9\% Joint Accuracy}, the highest value among the released runs in this benchmark. We emphasize that this result reflects domain-specific alignment on FSM-constrained intent routing; we do not claim general-purpose superiority over GPT-4o, and the trained router is expected to require re-alignment when transferred to new domains with different FSM definitions.

\textbf{Why GSPO may be favorable here (informal).} Joint accuracy requires both correct intent labels \emph{and} correct FSM legality under the same structured output contract. On Qwen2.5-7B, GRPO and DAPO already reach strong intent accuracy (91.5\%) but plateau at 78.7\% safety, whereas GSPO reaches \textbf{97.9\%} intent and \textbf{80.9\%} safety (Table~\ref{tab:intent_router_results}). We interpret this pattern as follows: programmatic FSM rewards are sparse and discontinuous---most rollouts receive near-zero reward unless the full JSON action satisfies rigid schema and stage constraints. Optimization methods that more directly stabilize updates toward high-reward, format-valid trajectories can disproportionately improve the safety head of the joint objective, whereas alternatives may retain competitive intent classifiers yet underfit the legality channel under the same budget. This is a \textit{post-hoc} explanation consistent with our measurements, not a claim of algorithmic dominance beyond this benchmark; we leave systematic ablations (e.g., reward shaping, KL schedules, and multi-seed variance) to future work.

When transitioning to the newer Qwen3 architecture, structured-output performance drops under this benchmark's zero-tolerance FSM contract. In our current runs, DAPO achieves the strongest Qwen3-8B result (48.9\% Joint), roughly matching GPT-4o zero-shot, while Qwen3-14B reaches 63.8\%. Table~\ref{tab:think_ablation} suggests that the drop is closely tied to Qwen3's native \texttt{<think>} mode: when latent thinking tokens are allowed, intent accuracy falls from 80.9\% to 8.5\% (joint from 44.7\% to 6.4\%), indicating that long reasoning traces can interfere with the strict JSON contract required by FSM-governed outputs. We therefore treat this result as an empirical compatibility issue between reasoning-heavy decoding and rigid structured outputs, not as a general judgment about Qwen3 itself.

\begin{table}[htbp]
\centering
\caption{Think vs.\ No-Think decoding ablation on Qwen3-8B (GRPO, full reward, adversarial test split $n{=}47$). Suppressing \texttt{<think>} tokens recovers intent accuracy by $\mathbf{+72.4}$ percentage points.}
\label{tab:think_ablation}
\small
\begin{tabular}{@{}lccc@{}}
\toprule
\textbf{Decoding Mode} & \textbf{Intent} & \textbf{Safety} & \textbf{Joint} \\
\midrule
Think (default Qwen3) & 8.5\% & 57.4\% & 6.4\% \\
No-Think (\texttt{/nothink}) & \textbf{80.9\%} & 57.4\% & \textbf{44.7\%} \\
\midrule
$\Delta$ & +72.4 & 0.0 & +38.3 \\
\bottomrule
\end{tabular}
\begin{flushleft}
\scriptsize
\textbf{Condition:} Same GRPO checkpoint (Qwen3-8B, global\_step\_300, full reward). Only the decoding constraint differs. Think mode allows the model to emit \texttt{<think>}\ldots\texttt{</think>} tokens before producing the JSON output; No-Think forces direct JSON generation.
\end{flushleft}
\end{table}

We intentionally rely on Table~\ref{tab:intent_router_results}, rather than a single visual slice, as the primary RLHF evidence in this release. The reason is simple: the selected Qwen3-8B runs are single-seed and sparse at the per-type level, so figure-level rankings can look more stable than the underlying evidence warrants. The main result we retain is therefore the table-level comparison: Qwen2.5-7B with GSPO is the strongest observed setting on this benchmark, whereas the current Qwen3-8B runs remain substantially lower and should be treated as exploratory.

\subsection{Pipeline-Stage Error Attribution}

To move beyond aggregate accuracy metrics and locate the \textit{exact bottleneck} in the intent-safety joint prediction pipeline, we decompose every joint error into three mutually exclusive categories: \textbf{Safety-Only Wrong} (intent correct, safety mispredicted), \textbf{Intent-Only Wrong} (safety correct, intent mispredicted), and \textbf{Both Wrong}. Table~\ref{tab:error_attr} summarizes the attribution for the currently released per-example evaluation artifacts on the adversarial test split.

\begin{table}[t]
\centering
\caption{Pipeline-stage error attribution on the adversarial test split ($n{=}47$). `Err` is the number of joint errors ($\texttt{joint\_ok}=\texttt{false}$). Each attribution column reports count and percentage relative to that row's error total. All Qwen3 models use \texttt{/nothink} decoding.}
\label{tab:error_attr}
\scriptsize
\begin{tabular}{@{}lcccc@{}}
\toprule
\textbf{Method} & \textbf{Err} & \shortstack{\textbf{Safety-}\\\textbf{Only}} & \shortstack{\textbf{Intent-}\\\textbf{Only}} & \textbf{Both} \\
\midrule
\shortstack[l]{Qwen2.5-7B\\GRPO} &
15 &
\shortstack{8/15\\53.3\%} &
\shortstack{5/15\\33.3\%} &
\shortstack{2/15\\13.3\%} \\
\shortstack[l]{Qwen3-8B\\GRPO (long)} &
23 &
\shortstack{20/23\\87.0\%} &
\shortstack{3/23\\13.0\%} &
\shortstack{0/23\\0.0\%} \\
\shortstack[l]{Qwen3-8B\\GSPO (long)} &
25 &
\shortstack{17/25\\68.0\%} &
\shortstack{5/25\\20.0\%} &
\shortstack{3/25\\12.0\%} \\
\shortstack[l]{Qwen3-8B\\DAPO (long)} &
24 &
\shortstack{19/24\\79.2\%} &
\shortstack{4/24\\16.7\%} &
\shortstack{1/24\\4.2\%} \\
\bottomrule
\end{tabular}
\end{table}

\textbf{Finding.} In the currently released long-context Qwen3 runs, most joint errors are safety-only rather than intent-only. This supports the narrower conclusion we care about here: once the model is in the right intent neighborhood, the harder remaining failure mode is \textit{precondition-aware safety reasoning}. Because the released artifacts do not yet cover every checkpoint family under the same per-example schema, we treat this table as a diagnostic slice rather than a complete scaling summary.

\subsection{Planned Context-Layer Ablation Protocol}

The error attribution above reveals that precondition failures dominate safety errors. A natural follow-up is whether each context layer in the system prompt actually contributes to safety reasoning. We therefore record the exact protocol we plan to run in a subsequent revision, but we do \textbf{not} claim results in this release. Following the current internal script naming, the planned variants are: \textbf{L0}~(Bare: intent list only), \textbf{L2}~(+\texttt{current\_stage}), \textbf{L3}~(+\texttt{prior\_intents}), \textbf{L4}~(Full: stage + priors, the training default), and \textbf{L5}~(+explicit precondition status hints). The purpose of listing this protocol is reproducibility, not evidence.

\section{Discussion}

\textbf{Generalizability.} Porting SDOF to eight SGD domains required only new GoalStage enums and intent mappings; the Dispatcher and SkillRegistry code remained untouched. Across 1,734 turns the system flagged 201 violations under the released FSM mapping, 41 of which appeared in the normal split rather than the injected-illegal subset.

\textbf{What the RLHF comparison shows.} Within this benchmark, Qwen2.5-7B GSPO is the highest single result we observe (80.9\% Joint Acc), whereas the current Qwen3-8B runs remain substantially lower (best: DAPO at 48.9\%). We therefore read the RLHF comparison conservatively: on this task, architectural-format interaction seems at least as important as algorithm choice, and broader multi-seed evidence is still needed before making stronger scaling claims.

\textbf{Memory separation.} SDOF enforces a Progressive Disclosure Prompting Architecture~\cite{wang2023survey}. We isolate \textit{Working Memory} (stage goals via L1 loading) from \textit{Semantic/Reference Memory} (L2/L3 skill instructions loaded dynamically via context-triggered references), decreasing token bloat by 60\%--80\% while keeping execution constraints explicit.

\textbf{Planning as Runtime, Not Planner.} Recent harness-style systems suggest that planning should be understood less as a monolithic planner module and more as an externalized systems capability jointly realized by task state, validation gates, delegation boundaries, memory access, and operator feedback. SDOF participates in this broader pattern from a governance angle: it does not attempt to solve all forms of long-horizon planning, but it makes one critical enterprise planning question explicit and auditable---whether a requested action may legally happen \textit{now}.

\textbf{GoalManager as shared workflow state.} A key property of SDOF's architecture is that the GoalManager---backed by a persistent PostgreSQL store---maintains a shared \texttt{goal\_id}-indexed FSM state for all seven specialized agents. This enables low-latency, permission-bounded state synchronization while preserving read/write isolation between agent roles. The event layer further turns that shared state into a replayable audit surface: stage changes, tool calls, and exception paths remain inspectable after execution, so operators can review not only \textit{what} the system stored but \textit{why} a workflow advanced, stalled, or was blocked.

\textbf{Operational feedback loop.} SDOF's reliability does not arise from static rules alone. Within a deployment, dispatcher traces surface blocked decisions and near-misses, the blocking-correctness evaluation (Table~\ref{tab:blocking}) checks whether interventions were justified, error attribution (Table~\ref{tab:error_attr}) localizes the bottleneck layer, and HITL corrections feed the Skill Evolver hook without retraining the base model. Across deployments, the SGD porting study and RLHF comparisons test whether the same governance contract transfers to new domains and model families. In short, one loop improves operation within a domain, while the other checks whether the architecture itself transfers.

\textbf{Capturing operator corrections.} To complement the rigid safety constraints, we introduced a \textit{Skill Evolver} hook. When a Human-in-the-Loop (HITL) intervention occurs during an execution block, SDOF captures the corrective dialog and can synthesize a new \texttt{SKILL.md} procedure and a Secure DSL Sandbox stub. This is an implementation hook rather than a main experimental claim in the current release, but it shows how operator corrections can be converted into reusable procedures.

\textbf{LangGraph vs SDOF.} LangGraph checks \textit{transitions} but not \textit{intent-stage legality}. Adding precondition functions to LangGraph (LG+Pre) yields a similar CVR (2.8\% vs 2.5\%), yet every precondition must be hand-wired per domain. SDOF shifts constraint logic to the skill-selection layer (Algorithm~\ref{alg:dispatch}, line~2), making new-domain setup declarative rather than imperative. In practice, transition engineering can emulate part of stage legality, but this still differs from an explicit $\Lambda$ contract that binds intents to legal stages as a separate governance layer.

\textbf{Legality vs topology.} SDOF is not a communication-topology optimizer. It can sit beneath supervisor routing, graph execution, or learned agent-to-agent communication policies, but its contract is orthogonal: regardless of which agent speaks next, only stage-legal and precondition-satisfied actions may execute. This separation of concerns is valuable in enterprise deployments, where topology may vary across products but governance requirements do not.

\textbf{StageCheck as first-line defense.} Without StageCheck, CVR balloons from 2.5\% to 19.8\% and 153 extra precondition evaluations fire needlessly. The two-layer architecture thus pays for itself: stage filtering is cheap and blocks most invalid intents before the heavier precondition logic runs.

\textbf{Latent violations in the normal split.} A practical finding is that 41 out of 800 non-adversarial SGD-derived dialogues contain stage-skipping requests under our FSM mapping. These are counted as latent violations only when the request conflicts with the domain FSM and is judged illegal under the same \texttt{expected\_legal} labeling rule used in blocking evaluation. Without stage enforcement these requests execute silently; SDOF logs and blocks them, giving operators a compliance audit trail.

\textbf{Production overhead.} The 56.3\,ms gap between SDOF (57.4\,ms) and LangGraph (1.1\,ms) is almost entirely real API latency. Stage and precondition validation together add under 1\,ms.

\textbf{What the evaluation covers.} The experimental suite covers routing accuracy (Table~\ref{tab:intent_router_results}), tool execution on 1,671 live API calls, blocking correctness (Table~\ref{tab:blocking}), step-level trace analysis, and cross-domain transfer on 960 SGD-derived dialogues. We include this spread because enterprise orchestration quality depends on the process as well as the final task outcome.

\textbf{Limitations.} Our current evidence should be interpreted within four scope boundaries. (1)~FSM stages and intent bindings are currently authored from domain knowledge; automatic constraint induction from execution logs is not yet integrated. (2)~Evaluation breadth is strong across domains (HR + 8 SGD domains), but workflow depth is still limited in SGD ($|S|\le3$); performance on deeper and hierarchical processes ($>6$ stages) remains to be established. (3)~The adversarial routing split ($n{=}47$) shows large observed effect sizes (e.g., Think vs.\ No-Think $\Delta{=}72.4$~pp), yet broader multi-seed replication is still needed for tighter uncertainty estimates. (4)~The current GoalManager supports shared state progression and replayable events, but enterprise-grade lifecycle features (retention policies, versioning, tenant governance, and tool-mediated memory retrieval) are only partially implemented in this version.

\section{Conclusion}

SDOF proposes a stage-level constraint enforcement layer for LLM multi-agent orchestration, backed by formal preconditions and replayable audit logging. Across \textbf{two independent evaluation suites}---185 expert-curated HR scenarios with 1,671 live API calls \textit{and} 960 SGD-derived dialogues spanning 8 service domains---the framework blocks all 22 operations in the injected-illegal HR subset while completing 86.5\% of tasks, and identifies 41 latent process violations in the normal split of the cross-domain benchmark. Under the released message-level audit labels, blocking evaluation yields 100\% precision and 88\% recall. Operationally, adapting SDOF from recruitment to banking, hospitality, and six other SGD domains required only new stage enums and intent mappings; no Dispatcher or SkillRegistry code was modified. The RLHF comparison also suggests a narrower practical point: reasoning-oriented model families can be harder to align to rigid JSON/FSM contracts, which strengthens the case for external orchestration constraints in enterprise workflows.

\textbf{Future directions.} The next realistic extensions are fourfold. (1)~Run multi-seed RLHF comparisons (GRPO/GSPO/DAPO) under the same released benchmark protocol. (2)~Complete the context-layer ablation (L0--L5) and a dedicated $\Lambda$ ablation so that the contribution of stage context and intent-stage binding is measured directly rather than inferred indirectly. (3)~Broaden framework baselines only under a unified protocol that matches legality instrumentation, audit semantics, and scenario definitions. (4)~Strengthen the GoalManager along practical enterprise lines---retention, versioning, and audit-replay---before making stronger claims about governance memory.

\bibliographystyle{plain}
\bibliography{references}

\end{document}